\pdfoutput=1

\documentclass[11pt]{article}

\usepackage[preprint]{acl}
\usepackage{times}
\usepackage{latexsym}
\usepackage{enumitem}
\usepackage{tabularx}
\usepackage{adjustbox}
\usepackage{multirow} 
\usepackage{xcolor}
\usepackage{colortbl}
\usepackage{mdframed} 
\usepackage{float}
\usepackage{tikz}
\usepackage{array}
\usepackage{placeins}
\usepackage{graphicx}
\usepackage{subcaption}
\usepackage{afterpage}
\usepackage{longtable}
\usepackage{booktabs} 
\usepackage{amsmath}
\usepackage{arydshln}

\usepackage[T1]{fontenc}

\usepackage[utf8]{inputenc}

\usepackage{microtype}

\usepackage{inconsolata}

\usepackage{graphicx}

\title{
GraphICL: Unlocking Graph Learning Potential in LLMs through Structured Prompt Design}

\author{
  Yuanfu Sun$^{1,2\dagger}$, Zhengnan Ma$^{1\dagger}$, Yi Fang$^{1}$, Jing Ma$^{3}$, Qiaoyu Tan$^{1*}$ \\
  $^1$Department of Computer Science, New York University (Shanghai) \\
  $^2$Courant Institute of Mathematical Sciences, New York University\\
  $^3$Department of Computer and Data Sciences, Case Western Reserve University \\
  \texttt{\{yuanfu.sun, zm2563, yf2722, qiaoyu.tan\}@nyu.edu,  jing.ma5@case.edu}\\
  \texttt{equal contribution$^\dagger$; corresponding author$^*$}
}

\begin{document}
\maketitle
\begin{abstract}
The growing importance of textual and relational systems has driven interest in enhancing large language models (LLMs) for graph-structured data, particularly Text-Attributed Graphs (TAGs), where samples are represented by textual descriptions interconnected by edges. While research has largely focused on developing specialized graph LLMs through task-specific instruction tuning, \textit{a comprehensive benchmark for evaluating LLMs solely through prompt design} remains surprisingly absent. Without such a carefully crafted evaluation benchmark, most if not all, tailored graph LLMs are compared against general LLMs using simplistic queries (e.g., zero-shot reasoning with LLaMA), which can potentially camouflage many advantages as well as unexpected predicaments of them. To achieve more general evaluations and unveil the true potential of LLMs for graph tasks, we introduce \textbf{Graph In-context Learning (GraphICL) Benchmark}, a comprehensive benchmark comprising novel prompt templates designed to capture graph structure and handle limited label knowledge. Our systematic evaluation shows that general-purpose LLMs equipped with our GraphICL outperform state-of-the-art specialized graph LLMs and graph neural network models in resource-constrained settings and out-of-domain tasks. These findings highlight the significant potential of prompt engineering to enhance LLM performance on graph learning tasks without training and offer a strong baseline for advancing research in graph LLMs. 
\end{abstract}

\section{Introduction}
Text-Attributed Graphs (TAGs), which integrate textual descriptions as node attributes with relational edges, form the foundation for understanding modern complex systems and relationships \citep{kipf2016semi,hamilton2017inductive}.  
Deep learning-based graph reasoning (GR) approaches, exemplified by graph neural networks (GNNs)~\citep{li2021training, sun2021scalable, zhou2020graph,tan2019deep,wu2022graph,reiser2022graph}, have achieved remarkable success in many TAG-related reasoning tasks, such as node classification ~\citep{fan2019graph,shi2023gigamae,wu2020comprehensive} and link prediction ~\citep{wu2022graph,reiser2022graph,tan2023s2gae,kipf2016variational}. 

However, most GNN-based approaches face two major hurdles: 
\tikz[baseline=(char.base)]{   \node[shape=circle,fill=black,text=white,inner sep=0.5pt] (char) {\textbf{\textit{1}}};
}
\textbf{Limited generalization across different graphs, particularly in cross-domain scenarios}. GNN models are typically tailored to specific graph structures they were originally trained on, and when applied to novel or cross-domain reasoning tasks, they exhibit a marked decline in performance~\citep{zhao2023graphtext,xu2024graphfm}. Resolving this often requires fine-tuning or full retraining, resulting in substantial computational overhead and deployment efforts. 
\tikz[baseline=(char.base)]{   \node[shape=circle,fill=black,text=white,inner sep=0.5pt] (char) {\textbf{\textit{2}}};
} \textbf{Performance depends heavily on labeled training graphs}. While GNNs perform well in supervised settings, their efficacy drastically diminishes in limited-label scenarios. Although graph few-shot learning~\citep{garcia2017few} has been introduced to mitigate this issue, it still requires a significant number of related learning tasks to adequately train the model for transfer to unseen tasks.    

To address these challenges, recent research has shifted from GNNs to graph LLMs~\citep{tang2023graphgpt,chenllaga,zhang2024graphtranslator,liu2024can,he2024unigraph,hu2024let, li2024llm}, most of them leverage LLMs' strong generalization capabilities for graph-related tasks through in-context learning (ICL)~\citep{dong2022survey}. Recent research on knowledge graph foundation models has also explored the idea of in-context learning for reasoning tasks \citep{cui2024prompt, galkin2023towards}. The key challenge for graph LLMs is incorporating graph structures into queries. Current approaches tackle this by either heuristically converting graphs into node sequences~\citep{chenllaga,ye2023natural} or embedding graph structures into hidden tokens via an auxiliary GNN~\citep{tang2023graphgpt,zhang2024graphtranslator,liu2024can,he2024unigraph}, which are then integrated into query templates for graph reasoning. By fine-tuning additional neural components or the general LLM backbones using graph-specific instruction tuning, these specialized methods have demonstrated superior zero-shot ICL capabilities compared to standard GNN studies. 

Despite the promising advances in specialized graph LLMs, their evaluation often relies on overly simplistic LLM baselines~\citep{tang2023graphgpt,chenllaga}, such as zero-shot reasoning with models like LLaMA or ChatGPT. Moreover, these models are typically assessed in in-domain scenarios and struggle to fully utilize limited labeled data (i.e., few-shot ICL), a capability that general-purpose LLM can readily support through effective prompt design. In the absence of a well-constructed evaluation benchmark, comparisons between specialized graph LLMs and general LLMs remain underexplored and poorly designed, potentially camouflaging many advantages as well unexpected predicaments of graph LLMs. This raises important questions: \textit{Can general-purpose LLMs be effectively adapted to tackle real-world graph reasoning tasks (e.g., node classification and link prediction) solely through in-context learning? Have we truly made progress in the development of graph LLMs?}

To address these questions, we propose \textbf{GraphICL}: Benchmarking Large Language Models for \textbf{Graph} Reasoning
via \textbf{I}n-\textbf{C}ontext \textbf{L}earning. In our framework, GraphICL refers to the design of task-specific prompts following a unified GraphICL template across diverse graph reasoning tasks. 
GraphICL facilitates graph reasoning in LLMs by leveraging four core components: task description, anchor node text, structure-aware information, and labeled demonstrations. By incorporating anchor nodes and their $k$-hop neighbors, we enable zero-shot graph reasoning, utilizing the inherent relationships between proximate nodes. Through strategical selection of neighbors and demonstrations, such as the top $M$ most similar or influential nodes, we optimize few-shot reasoning, releasing the potential of LLMs. GraphICL pushes the boundaries of LLMs' capabilities in graph tasks, enabling performance that was previously unattainable. Our key contributions are summarized as follows:
\begin{itemize}[leftmargin=*, topsep=0mm]
    \item \textbf{Novel Research Problem.} 
    We investigate whether better graph reasoning (GR) results can be achieved by simply prompting LLMs through GraphICL, without additional training, and whether this approach can outperform both supervised GNNs and specialized Graph LLMs in both in- and cross-domain scenarios. 
    
    \item \textbf{A Comprehensive Prompt Benchmark for LLM in Graph Reasoning.} 
    Previous comparisons between general LLMs and specialized graph models have been biased by underdeveloped prompts, which fail to harness the full potential of LLMs. We propose GraphICL, a comprehensive prompt set that encompasses graph structure, labeled demonstration, and diverse evaluation tasks. 
    \item \textbf{Systematic Evaluation.} 
    We conducted extensive experiments on 9 datasets, encompassing both in-domain and cross-domain scenarios, and benchmarked our approach against state-of-the-art graph LLMs as well as traditional supervised GNN models. Additionally, we performed comprehensive ablation studies to assess the impact of various prompt configurations within the GraphICL framework.
    \item \textbf{Promising Observations.} 
    Our extensive evaluation yielded several valuable insights that can inform the future application of LLMs in graph reasoning, particularly through in-context learning. These findings also establish a solid foundation for advancing research in graph LLMs. 
\end{itemize}

\section{Related Work}
\noindent\textbf{Specialized Graph LLMs.} Building on the success of large language models (LLMs), the application of LLMs to graph reasoning tasks has gained considerable attention. The core idea is to incorporate graph structures into queries and then instruction-tune the LLMs using graph-related tasks. Based on graph transformation strategies, existing efforts can be broadly categorized into two approaches: heuristic and learnable. The heuristic approach~\citep{chenllaga,liu2024moleculargpt,ye2023natural} converts graphs into node sequences using manually designed transformation rules~\citep{ye2023natural} or random walks~\citep{chenllaga}. In contrast, the learnable approach~\citep{tang2023graphgpt,zhang2024graphtranslator,liu2024can,he2024unigraph} encodes graph structures into hidden sequences through additional GNN encoders, which are either pre-trained~\citep{he2024unigraph,fang2024uniglm,tang2023graphgpt} or jointly fine-tuned~\citep{zhang2024graphtranslator,liu2024can} with the LLM backbone during instruction tuning. While these specialized graph LLMs inherit the zero-shot in-context learning (ICL) capabilities of general LLMs, they struggle to fully utilize few-shot demonstrations for performing few-shot ICL on graphs. This limitation hinders their ability to adapt effectively to tasks requiring additional contextual information.

\noindent\textbf{General-purpose LLM for Graph Reasoning.} In parallel, another line of research represents graph structures using natural language descriptions, combining them with task-specific templates to query general-purpose LLMs. Notable works such as \citep{huang2023can,zhao2023graphtext,guo2023gpt4graph,chen2024exploring,li2024similarity,shi2024retrieval,fang2024gaugllm} have advanced this area, primarily focusing on using LLMs for graph augmentations~\citep{he2023explanations,chen2024exploring,fang2024gaugllm}. While some efforts~\citep{huang2023can} have explored graph structure's role in LLM inference through both zero-shot and few-shot ICL, they remain limited in terms of prompt template diversity, neighborhood and labeled demonstration selection, evaluation scenarios, and the breadth of GR tasks.

In contrast, we introduce a comprehensive prompt template design for graph reasoning tasks, where the prompts in~\citep{huang2023can,li2024similarity} can be seen as a subset of our approach. More importantly, we benchmark the performance of specialized graph LLMs and general-purpose LLMs equipped with our prompt suite, offering a timely and fair comparison of recent specialized graph LLM studies while providing insights into their strengths and weaknesses relative to general LLMs utilizing prompt design. 


\section{Problem Statement}

Given a Text-Attributed Graph \( \mathcal{G} = (\mathcal{V}, \mathcal{A}, \mathcal{T}) \), where \( \mathcal{V} \) represents nodes, \( \mathcal{A} \) is the adjacency matrix, and \( \mathcal{T} \) contains the node texts, along with a LLM $f(\cdot)$, this paper aims to leverage Graph In-Context Learning (GICL) to generate relevant GICL-Prompts $P$, which are the outputs of $GP(\cdot)$, as inputs for the LLM $f(\cdot)$ to solve two classic graph reasoning tasks: node classification (NC) and link prediction (LP).

\noindent\textbf{Node Classification via GICL.}
For node classification, we can use two different GICL methods to predict the label \( y_i \) of node \( \mathcal{V}_i \) in Graph \( \mathcal{G} \).

i) \textit{\textbf{NC-Zero-shot:}} Use only the anchor node text \( \mathcal{T}_i \), or include neighboring node texts \( \mathcal{T'} \), as the main content to generate the GICL-Prompt, where \( P = GP(\mathcal{T}_i, \mathcal{T'}) \). This prompt is then fed into the LLM to obtain the prediction, \( y_p = f(P) \).

ii) \textit{\textbf{NC-Few-shot:}} Building upon the zero-shot template, we further incorporate neighboring nodes' texts \( \mathcal{T'} \) and their labels \( \mathcal{Y'} \), or additionally include demonstration texts \( \mathcal{T''} \) and labels \( \mathcal{Y''} \), to form a more informative GICL-Prompt, where \( P = GP(\mathcal{T}_i, \mathcal{T'}, \mathcal{Y'}, \mathcal{T''}, \mathcal{Y''}) \). This enriched prompt is then input into the LLM to generate the final prediction, \( y_p = f(P) \).

\noindent\textbf{Link Prediction via GICL.}
For link prediction between nodes \( \mathcal{V}_m \) and \( \mathcal{V}_n \), we can also utilize these two approaches:

i) \textit{\textbf{LP-Zero-shot:}} We begin by using the textual information of the two nodes, \( \mathcal{T}_m \) and \( \mathcal{T}_n \), and optionally incorporate neighboring node texts \( \mathcal{T'} \) to construct a GICL-Prompt, \( P = GP(\mathcal{T}_m, \mathcal{T}_n, \mathcal{T'}) \). This prompt is then passed into the LLM to predict the existence of a link, \( y_p = f(P) \).

ii) \textit{\textbf{LP-Few-shot:}} To further improve performance, we introduce demonstration texts \( \mathcal{T''} \) and corresponding link relationships to enrich the Prompt, \( P = GP(\mathcal{T}_m, \mathcal{T}_n, \mathcal{T'}, \mathcal{T''}) \). This more comprehensive prompt is then used by the LLM to generate a refined link prediction, \( y_p = f(P) \).

\begin{figure*}[ht]
  \centering
  \includegraphics[width=0.98\linewidth]{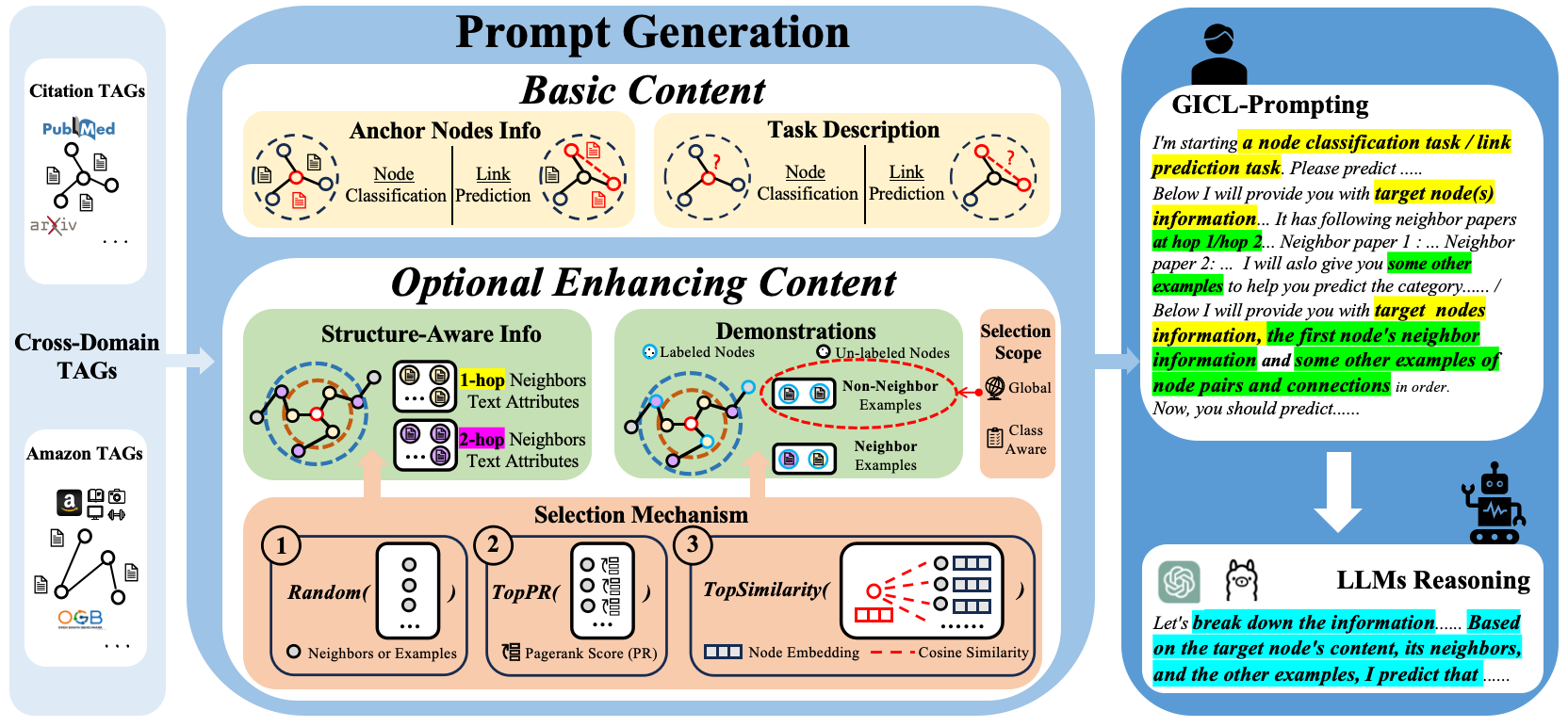}
  \caption {The overall framework of our GraphICL. We implement various graph in-context learning templates by combining basic content with optional enhancing content. These templates are then input as prompts into large language models to obtain relevant prediction results.}
  \label{roadmap}
\end{figure*}

\section{Prompt Design Driven by Graph In-Context Learning} 

In this section, we will explain how each type of graph in-context learning method is implemented within our framework. Our GraphICL prompt template consists of 4 fundamental components: \textit{anchor node text}, \textit{task description}, \textit{structure-aware information}, and \textit{demonstrations}, as shown in Figure~\ref{roadmap}. By combining these 4 components, we can generate 55 different prompt templates. These templates are applied to 2 classic graph reasoning tasks on 9 datasets. By comparing with multiple models, we demonstrate the significant boost our GICL template provides to various of LLMs. Section ~\ref{sec:3.1} and ~\ref{sec:3.2} explain the generation and function of each component of the prompt respectively. Section ~\ref{sec:3.3} shows how different modules of the prompt are combined to form the final input for the LLMs.

\subsection{Basic Content}
\label{sec:3.1}
The basic content primarily conveys the information specific to the anchor node, ensuring that the LLM comprehends the graph reasoning task it is expected to execute. It constitutes a critical component of the general prompt and serves as the foundational text in GraphICL.

\noindent\textbf{Text of Anchor Nodes.}
The text associated with the anchor node can vary, such as the title and abstract of a paper \citep{shibata2012link} or the description of a product \citep{hasan2011survey}. In the context of link prediction, however, the anchor nodes refer to both the source and target nodes of the predicted edge. The corresponding text in this case is the concatenated text of these two nodes.

\noindent\textbf{Task Description.}
For different graph reasoning tasks, it is crucial to explicitly define the task objectives for the LLMs. This guiding piece of text is referred to as the task description. In node classification tasks, for instance, LLMs may not inherently recognize the specific names of categories within the dataset. Therefore, we explicitly provide the names of all labels within the task description. Typically, the task description serves as the system prompt \citep{giray2023prompt} for LLMs.

\subsection{Optional Enhancing Content} 
\label{sec:3.2}
While the basic content provides the essential information needed for LLMs to perform fundamental graph zero-shot learning, it alone is insufficient. To enhance the LLM's ability to reason effectively, additional structural information and other relevant data must be integrated. This supplementary layer of information, known as \textbf{Enhancing Content}, serves to deepen the LLMs' understanding and reasoning capabilities.

\noindent\textbf{Structure-Aware Information.} 
Graph structures exhibit complex dependencies, prompting GNNs to employ message passing for gathering and updating node information from neighbors, enriching node representations \citep{zhou2020graph}. In graph in-context learning, we simulate this by providing textual information from an anchor node’s neighbors for a well-established LLM, effectively enabling message passing at the textual level. We focus on 1-hop neighbors for capturing immediate, direct influences on the target node, representing short-term dependencies, and 2-hop neighbors due to their importance in graph reasoning, as GNNs typically utilize two layers \citep{chen2024exploring}. By incorporating 1-hop or 2-hop neighbors' text, our framework enriches structure-aware information and enhances the flexibility of prompt design.

The strategy for selecting neighbors is also crucial, as there is typically no one-size-fits-all approach that achieves optimal results across all graph reasoning tasks. Below, we will introduce three selection strategies employed in our work: 

(1) \textbf{Random Selection:} Randomly selecting $M$ nodes from the $k$-hop neighbors of the anchor node, treating each neighbor as equally contributing to the reasoning process. 

(2) \textbf{Similarity-based Selection:} Calculating cosine similarity between the anchor node and $k$-hop neighbors, selecting the Top $M$ most similar neighbors, prioritizing those with higher textual relevance to the anchor node. 

(3) \textbf{PageRank-based Selection:} Computing PageRank scores \citep{page1999pagerank} for each $k$-hop neighbor, selecting the Top $M$ based on their importance within the graph structure.

\noindent\textbf{Few-shot Demonstrations.} 
Demonstrations are a crucial component of few-shot learning for LLMs, with their design tailored to specific downstream tasks. These demonstrations are intended to aid LLMs in gaining a deeper understanding of the tasks and enhancing their reasoning abilities \citep{brown2020language}. For tasks such as node classification and link prediction, additional text from other nodes, along with their corresponding labels, must be provided to facilitate the LLMs' inference.

When selecting demonstrations, we employ the same three strategies used for neighbor selection: random selection, similarity-based selection, and PageRank-based selection. The selection scope for demonstrations can be either "Global", where $M$ demonstrations are chosen from the training set using these selection methods, or "Class-Aware", where one demonstration is provided for each class label. In the latter case, the selection method for each demonstration of each label follows one of the aforementioned three strategies. 

\subsection{Graph In-Context Learning Prompting}
\label{sec:3.3}
This section discusses how to integrate the four components of the graph in-context learning prompt for different graph reasoning tasks. 
Examples are provided in Figure~\ref{figure2} for further illustration.

\noindent\textbf{NC-Graph Zero-shot Prompting:} The zero-shot prompt includes the Basic Content, which consists of the anchor node’s text and a description of the classification task. Structure-aware information can be optionally added as enhanced content but without including labels of $k$-hop neighbors.

\noindent\textbf{NC-Graph Few-shot Prompting:} Similar to zero-shot, the few-shot prompt also includes Basic Content.Additionally, it provides non-neighbor labeled demonstrations as enhanced content; moreover, the enhanced content can also include neighbor information with labels. Alternatively, labeled neighbor information can also be used as a demonstration for few-shot learning, where structure-aware information is omitted to avoid redundancy.

\noindent\textbf{LP-Graph Zero-shot Prompting:} Providing the textual information of both the start and end nodes of the target relationship, with the option to include neighbor information for one of the nodes. 

\noindent\textbf{LP-Graph Few-shot Prompting:} Building on zero-shot template, it adds connection relationships between node pairs from the training set, along with their textual information. The selection of these relationships follows the three methods outlined in section~\ref{sec:3.2}. Here, the PageRank score of an edge is defined as the average PageRank of its two end nodes, while the edge embedding is computed as the average of their embeddings. Unlike in node classification, the relationships between a node and its neighbors cannot be directly used as demonstrations, as they are implicitly encoded within the structure-aware information. 

\section{Experiments}
In this section, we present a rigorous evaluation of specialized graph LLMs and general-purpose LLMs equipped with our GraphICL. The experiments are divided into two key parts: a comparative analysis of GraphICL against state-of-the-art graph LLMs across various scenarios, and an exploration of how different GraphICL configurations impact the performance of general-purpose LLMs. Addtional details (hyperparameter settings and results) can be found in Appendix~\ref{sec:A.3}.

\begin{table*}[t]
    \caption{In-domain node classification results: For MLP, GCN \citep{kipf2016semi}, RevGAT \citep{li2021training}, and SAGE \citep{sun2021scalable}, we uniformly use BERT embeddings \citep{devlin2018bert}. Acc(\%) is used as the evaluation metric, and we calculate the relative difference between the best results of our method and others. For each LLaMA model in GraphICL, S1 and S2 denote the first- and second-best GraphICL prompts. "NA" indicates that the result is unavailable. The results in 
    \textbf{\textcolor{blue}{blue}} and \textbf{\textcolor{red}{red}} respectively represent the best baseline results under the semi-supervised and supervised settings. For further explanation and settings, please refer to the Appendix~\ref{sec:A.4}.}
    \label{tab:Table1}
    \centering
    \scriptsize
    \renewcommand{\arraystretch}{1.4} 
    \begin{tabularx}{\textwidth}{|c|X|X|X|X|X|X|}
        \hline
        \rowcolor{blue!10} \textbf{Method} & \textbf{Computers} & \textbf{Sports} & \textbf{PubMed} & \textbf{Cora} & \textbf{Arxiv} & \textbf{Products} \\
        \hline
        \multicolumn{7}{>{\columncolor{gray!20}}c}{\textbf{Semi-Supervised In-Domain Results}} \\
        \hline
        MLP & 44.56 (+96.07\%) & 58.74 (+55.58\%) & 59.38 (+56.92\%) & 47.23 (+76.96\%) & 37.10 (+98.60\%) & 65.36 (+24.66\%) \\
        GCN & 59.12 (+47.78\%) & 70.24 (+30.11\%) & 74.25 (+25.49\%) & \textbf{\textcolor{blue}{68.82 (+21.45\%)}} & \textbf{\textcolor{blue}{55.27 (+33.31\%)}} & 74.47 (+9.41\%) \\
        SAGE & 58.52 (+49.30\%) & 69.53 (+31.44\%) & 64.66 (+44.11\%) & 64.58 (+29.42\%) & 54.05 (+36.32\%) & 72.35 (+12.62\%) \\
        RevGAT & 55.48 (+57.48\%) & 64.63 (+41.40\%) & 64.10 (+45.37\%) & 65.31 (+27.97\%) & 48.86 (+50.80\%) & 71.45 (+14.04\%) \\
        LLaGA-ND & 49.48 (+76.58\%) & 52.19 (+75.11\%) & 39.96 (+133.18\%) & 48.52 (+72.26\%) & 54.26 (+35.79\%) & 73.32 (+11.13\%) \\
        LLaGA-HO & 55.68 (+56.91\%) & 63.81 (+43.22\%) & 40.37 (+130.81\%) & 40.96 (+104.05\%) & 53.02 (+38.97\%) & 72.76 (+11.98\%) \\
        GraphGPT & NA & NA & NA & NA & NA & NA \\
        GraphPrompter &\textbf{ \textcolor{blue}{62.46 (+39.88\%)}} & \textbf{\textcolor{blue}{80.92 (+12.94\%)}} & \textbf{\textcolor{blue}{88.11 (+5.75\%)}} & 51.11 (+63.53\%) & 54.12 (+36.14\%) & \textbf{\textcolor{blue}{76.34 (+6.73\%)}} \\
        GraphTranslator & 38.95 (+124.31\%) & 22.88 (+299.43\%) & 60.46 (+54.12\%) & 35.59 (+134.84\%) & 28.48 (+158.71\%) & 41.32 (+97.19\%) \\
        \hline
        \multicolumn{7}{>{\columncolor{gray!20}}c}{\textbf{Supervised In-Domain Results}} \\
        \hline
        MLP & 61.74 (+41.51\%) & 85.06 (+7.44\%) & 82.55 (+12.88\%) & 63.12 (+32.41\%) & 69.20 (+6.47\%) & 67.56  (+20.60\%) \\
        GCN & 74.35 (+17.51\%) & 88.14 (+3.69\%) & 86.43 (+7.81\%) & 75.08 (+11.32\%) & 72.63 (+1.45\%) & 75.56 (+7.83\%) \\
        SAGE & 73.54 (+18.81\%) & 89.00 (+2.69\%) & 86.26 (+8.02\%) & 74.94 (+11.53\%) & 73.33 (+0.48\%) & 73.44 (+10.94\%) \\
        RevGAT & 73.16 (+19.42\%) & 87.18 (+4.83\%) & 86.70 (+7.47\%) & 74.21 (+12.63\%) & 72.88 (+1.10\%) & 73.62 (+10.67\%) \\
        LLaGA-ND & \textbf{\textcolor{red}{86.99 (+0.44\%)}} & 90.91 (+0.53\%) & 88.89 (+4.83\%) & 88.19 (-5.23\%) & 73.04 (+0.88\%) & 73.62 (+10.68\%) \\
        LLaGA-HO & 78.78 (+10.90\%) & 88.35 (+3.44\%) & 88.77 (+4.97\%) & \textbf{\textcolor{red}{88.56 (-5.62\%)}} & \textbf{\textcolor{red}{74.02 (-0.46\%)}} & 73.64 (+10.65\%) \\
        GraphGPT & NA & NA & 84.68 (+10.04\%) & NA & 62.18 (+18.49\%) & NA \\
        GraphPrompter & 78.38 (+11.47\%) & \textbf{\textcolor{red}{91.85 (-0.50\%)}} & \textbf{\textcolor{red}{94.32 (-1.21\%)}} & 70.11 (+19.21\%) & 72.38 (+1.80\%) & \textbf{\textcolor{red}{79.04 (+3.09\%)}} \\
        GraphTranslator & 38.95 (+124.31\%) & 22.88 (+299.43\%) & 60.46 (+54.12\%) & 35.59 (+134.84\%) & 28.48 (+158.71\%) & 41.32 (+97.19\%) \\
        \hline
        \multicolumn{7}{>{\columncolor{gray!20}}c}{\textbf{Results of Zero-shot LLMs}} \\
        \hline
        \arrayrulecolor{black}
        LLaMA3-70b-Instruct & 57.42 (+52.16\%) & 67.45 (+35.49\%) & 91.94 (+1.35\%) & 66.24 (+26.18\%) & 62.72 (+17.47\%) & 57.62 (+41.41\%) \\
        LLaMA2-13b-Chat & 47.48 (+84.01\%) & 35.39 (+158.24\%) & 74.04 (+25.85\%) & 52.58 (+58.96\%) & 44.04 (+67.30\%) & 57.34 (+42.10\%) \\
        \hline
        \multicolumn{7}{>{\columncolor{orange!15}}c}{\textbf{Results of GraphICL (Ours)}} \\
        \hline
        \arrayrulecolor{orange}
        GraphICL-LLaMA3-S1 & \textbf{87.37} & \textbf{91.39} & \textbf{93.18} & \textbf{83.58} & \textbf{73.68} & \textbf{81.48} \\
        GraphICL-LLaMA3-S2 & 87.37 & 91.12 & 93.05 & 83.21 & 73.54 & 81.04 \\
        GraphICL-LLaMA2-S1 & 87.06 & 85.52 & 82.56 & 77.49 & 70.20 & 78.84 \\
        GraphICL-LLaMA2-S2 & 85.11 & 83.19 & 79.54 & 76.94 & 69.82 & 78.27 \\
        \arrayrulecolor{black}
        \hline
    \end{tabularx}
\end{table*}

\subsection{Experiment Configurations}
\label{sec:4.1}
\textbf{Datasets.} We conducted experiments on two major types of datasets: Citation Networks and Amazon Review Datasets. The Citation Networks include PubMed \citep{sen2008collective}, Cora \citep{mccallum2000automating}, and OGB-Arxiv \citep{hu2020open}, while the Amazon review datasets include OGB-Products \citep{hu2020open}, Amazon-Photo, Amazon-History, Amazon-Computers, Amazon-Sports-Fitness and Amazon-Children-Book \citep{shchur2018pitfalls}. 
For specific data splits, please refer to Appendix~\ref{sec:A.3.1}. For the results of Amazon-Photo, Amazon-History and Amazon-Children-Book, please refer to Table~\ref{tab:extra in-domain results} and Table~\ref{tab:extra_cross_domain} in Appendix. 
    
\noindent \textbf{Large Language Models.} We utilized a total of three language models for testing: LLaMA2-13B-Chat\footnote{https://llama.meta.com/llama2/}, LLaMA3-70B-Instruct\footnote{https://llama.meta.com/llama3/}, and GPT-4o\footnote{https://platform.openai.com/docs/models/gpt-4o}. Due to budget constraints, we did not conduct comprehensive experiments with GPT-4o.

\noindent \textbf{Baselines.} 
In our performance evaluation, we take into account a range of state-of-the-art methods for a thorough assessment. (i) The first category consists of MLP, which utilizes a Multilayer Perception for prediction. (ii) The second category includes prominent GNN encoders, such as GraphSAGE \citep{li2021training}, GCN \citep{kipf2016semi}, RevGAT \citep{li2021training}. (iii) The third category encompasses influential specialized Graph LLMs, including LLaGA (ND, HO) \citep{chen2024llaga}, GraphGPT \citep{tang2023graphgpt}, GraphTranslator \citep{zhang2024graphtranslator}, and GraphPrompter \cite{liu2024can}. (iv) The last category consists of pure zero-shot LLMs, which can also be viewed as methods that input basic content into LLMs for reasoning. In our experiments, all the settings used for GraphICL are explained in detail in the Appendix~\ref{sec:A.2}.

\subsection{Multi-scenario Graph Reasoning Testing}
\label{sec:4.2}

We begin by analyzing the node classification results, focusing on two scenarios: \textbf{in-domain} and \textbf{cross-domain}. In the in-domain scenario, testing is performed on datasets used during training, whereas in the cross-domain scenario, the test datasets have no overlap with the training data.

\subsubsection{In-Domain Node Classification}

\textbf{RQ 1.} 
\textit{Can GICL-prompted LLMs outperform state-of-the-art GNNs and specialized GraphLLMs in the in-domain scenario?}

\noindent \textbf{Experiment Settings.}
We used six datasets (see Table~\ref{tab:Table1}) to evaluate our GraphICL method combined with LLaMA2 and 3 (as described in Section ~\ref{sec:4.1}), comparing it against various GNNs, Graph LLMs, and LLM methods. For LLM-based methods, only the single most likely label was predicted, and accuracy was calculated accordingly. 

\noindent \textbf{\underline{Observation 1.}} 
\textit{\textbf{Equipped with GraphICL, general-purpose LLMs can achieve competitive or even superior performance compared to specialized graph LLMs in both semi-supervised and supervised settings in the in-domain scenario.}} Specifically, in the semi-supervised setting, GraphICL achieves an average relative improvement of around 20\% across datasets, with a significant 39.88\% increase on the Computers dataset compared to GraphPrompter, showcasing its robust performance. Even in the supervised setting, GraphICL continues to outperform most graph LLMs and all GNNs, consistently demonstrating its superiority. Moreover, it exhibits a marked improvement in reasoning capabilities over zero-shot LLMs, further solidifying its effectiveness and adaptability across different learning paradigms and graph reasoning tasks.

\begin{table*}[t]
    \caption{Cross-Domain results of node classification. In this setting, none of the Graph LLM methods were trained or fine-tuned on the training set of the corresponding dataset being tested. Below, "NA" indicates "Not Applicable," meaning the corresponding dataset is part of the training set. The results in 
     \textbf{\textcolor{red}{red}} represent the best baseline results.}
    \label{tab:out_graph}
    \centering
    \scriptsize
    \renewcommand{\arraystretch}{1.4} 
    \begin{tabularx}{\textwidth}{|c|X|X|X|X|X|X|}
        \hline
        \rowcolor{blue!10} \textbf{Method} & \textbf{Computers} & \textbf{Sports} & \textbf{PubMed} & \textbf{Cora} & \textbf{Arxiv} & \textbf{Products} \\
        \hline
        \multicolumn{7}{>{\columncolor{gray!20}}c}{\textbf{Supervised Cross-Domain Results (GraphLLMs)}} \\
        \hline
        LLaGA-ND & 14.88 (+487.16\%) & 3.57 (+2459.94\%) & NA & NA & NA & NA \\
        LLaGA-HO & 14.71 (+493.95\%) & 4.84 (+1788.22\%) & NA & NA & NA & NA \\
        GraphGPT & 14.61 (+498.02\%) & 8.24 (+1009.10\%) & NA & 41.14 (+103.16\%) & NA & 31.67 (+157.28\%) \\
        GraphPrompter & 26.40 (+230.95\%) & 9.26 (+886.93\%) & NA & NA & 3.62 (+1935.36\%) & 15.42 (+428.40\%) \\
        GraphTranslator & 32.85 (+165.97\%) & 12.9 (+608.44\%) & 46.17 (+101.82\%) & 34.06 (+145.39\%) & NA & 18.31 (+345.00\%) \\
        \hline
        \multicolumn{7}{>{\columncolor{gray!20}}c}{\textbf{Results of Zero-shot LLMs}} \\
        \hline
        \arrayrulecolor{black}
        LLaMA3-70b-Instruct & \textbf{\textcolor{red}{57.42 (+52.16\%)}} & \textbf{\textcolor{red}{67.45 (+35.49\%)}} & \textbf{\textcolor{red}{91.94 (+1.35\%)}} & \textbf{\textcolor{red}{66.24 (+26.18\%)}} & \textbf{\textcolor{red}{62.72 (+17.47\%)}} & \textbf{\textcolor{red}{57.62 (+41.41\%)}} \\
        LLaMA2-13b-Chat & 47.48 (+84.01\%) & 35.39 (+158.24\%) & 74.04 (+25.85\%) & 52.58 (+58.96\%) & 44.04 (+67.30\%) & 57.34 (+42.10\%) \\
        \hline
        \multicolumn{7}{>{\columncolor{orange!15}}c}{\textbf{Results of GraphICL (Ours)}} \\
        \hline
        \arrayrulecolor{orange}
        GraphICL-LLaMA3-S1 & \textbf{87.37} & \textbf{91.39} & \textbf{93.18} & \textbf{83.58} & \textbf{73.68} & \textbf{81.48} \\
        GraphICL-LLaMA3-S2 & 87.37 & 91.12 & 93.05 & 83.21 & 73.54 & 81.04 \\
        GraphICL-LLaMA2-S1 & 87.06 & 85.52 & 82.56 & 77.49 & 70.20 & 78.84 \\
        GraphICL-LLaMA2-S2 & 85.11 & 83.19 & 79.54 & 76.94 & 69.82 & 78.27 \\
        \arrayrulecolor{black}
        \hline
    \end{tabularx}
    \vspace{-0.4cm}
\end{table*}

\subsubsection{Cross-Domain Node Classification}
\textbf{RQ 2.} 
\textit{Can GICL-prompted LLMs excel over top GNNs and GraphLLMs in Cross-Domain tasks with mismatched training and testing data?}

\noindent \textbf{Experiment Settings.}
We used the same six datasets as the in-domain testing phase. Given that GNNs lack robust cross-domain capabilities, this experiment focused on directly comparing GraphICL with tailored and specialized GraphLLMs.

\noindent \textbf{\underline{Observation 2.}} 
\textit{\textbf{In the cross-domain scenario, GraphICL enables LLaMA to outperform tailored Graph LLMs without requiring additional training,}} demonstrating a significant advantage. For the Graph LLM methods, we employed a diverse combination of mixed training sets to enhance their cross-domain capabilities. However, despite these efforts, both Graph LLM and zero-shot LLM methods fall considerably short, with the former showing a relative performance gap exceeding 101\%, showcasing its potential to adapt LLMs to unseen graph data and broader applications.

\subsubsection{Link Prediction Testing}
For link prediction, the substantial increase in text data significantly extends the testing time, making it impractical to perform exhaustive evaluations across all datasets. Therefore, we selected Cora for multi-scenario testing, similar to the approach used for node classification, to maintain consistency and ensure a thorough evaluation. As shown in Table~\ref{tab:link_prediction}, our GraphICL method consistently achieves the best performance compared to other models. Notably, in the supervised setting, it outperforms the best result from the remaining methods, including LLaGA-HO, by 1.26\%, highlighting its robustness. This further confirms the observations made in the node classification task, showcasing GraphICL's superior generalization and reasoning capabilities across various graph-related tasks.


\begin{table}[ht]
\centering
\footnotesize
\caption{Link prediction results in Cora. For GCN and GraphSAGE, we use sbert embeddings \cite{reimers2019sentence}.}
\begin{tabular}{|c|c|c|}
\hline
\textbf{Train → Test} & \textbf{Method} & \textbf{Accuracy} \\ \hline
\multirow{8}{*}{\begin{tabular}[c]{@{}c@{}}Cora (Semi-Supervised)\\ ↓\\ Cora\end{tabular}} 
& GCN            & 58.97  \\ 
& GraphSAGE      & 67.68  \\ 
& GraphGPT    & -  \\ 
& LLaGA-ND    & 58.38  \\ 
& LLaGA-HO    & 59.12  \\
& LLaMA2    & 75.00  \\ 
& LLaMA3    & 84.11  \\
& GraphICL (Ours)    & \textbf{88.08}  \\ \hline
\multirow{8}{*}{\begin{tabular}[c]{@{}c@{}}Cora (Supervised) \\ ↓\\ Cora\end{tabular}} 
& GCN            & 81.59  \\ 
& GraphSAGE      & 79.15  \\ 
& GraphGPT    & 80.26  \\ 
& LLaGA-ND    & 83.79  \\ 
& LLaGA-HO    & 86.82  \\
& LLaMA2    & 75.00  \\ 
& LLaMA3    & 84.11  \\
& GraphICL (Ours)    & \textbf{88.08}  \\ \hline
\multirow{8}{*}{\begin{tabular}[c]{@{}c@{}}Arxiv+PubMed\\ ↓\\ Cora\end{tabular}} 
& GCN            & 56.73  \\ 
& GraphSAGE      & 58.92  \\ 
& GraphGPT    & 50.74  \\ 
& LLaGA-ND   & 86.47  \\ 
& LLaGA-HO    & 87.35  \\
& LLaMA2    & 75.00  \\ 
& LLaMA3    & 84.11  \\
& GraphICL (Ours)    & \textbf{88.08}  \\ \hline
\end{tabular}
\vspace{-0.4cm}
\label{tab:link_prediction}
\end{table}

\subsection{Impact of GraphICL Configuration}
Factors such as the type of LLMs using the GICL method and the inclusion of structural information can affect performance. In this section, we will explore these main influencing factors.
\subsubsection{LLMs Comparison with GraphICL }
\label{sec:codl}

\begin{table*}[th]
    \caption{The node classification accuracies of different LLMs under several different GraphICL methods on two dataset. $S-A$ represents specific neighborhood information in structure-aware, $S_{SM}$ denotes the neighbor selection method, $Demo$ denotes the type of demonstrations, and $D_{SM}$ denotes the method of selecting demonstrations.}
    \label{tab:different_llm}
    \footnotesize
    \begin{tabularx}{\textwidth}{c c*{5}{X}}
        \toprule
        \multirow{2}{*}{Dataset} & \multirow{2}{*}{LLM} & \multicolumn{4}{c}{GraphICL} & \multirow{2}{*}{Acc(\%)} \\ \cmidrule(lr){3-6}
        & & $S-A$ & $S_{SM}$ & $Demo$ & $D_{SM}$ \\ 
        \midrule
        \multirow{3}{*}{Cora} 
        & GPT-4o & 1-hop & Similarity & global & Pagerank & 76.60 \\
        & LLaMA3-70B-Instruct & 1-hop & Similarity & global & Pagerank & 75.40 \\
        & LLaMA2-13B-Chat & 1-hop & Similarity & global & Pagerank & 70.60 \\
        \midrule
        \multirow{3}{*}{Sports} 
        & GPT-4o & 1-hop & Random & global & Random & 91.00 \\
        & LLaMA3-70B-Instruct & 1-hop & Pagerank & global & Random & 84.80 \\
        & LLaMA2-13B-Chat & 1-hop & Pagerank & global & Random & 75.90 \\
        \bottomrule
    \end{tabularx}
    \vspace{-0.4cm}
\end{table*}

\textbf{RQ 3.} 
\textit{How does the performance vary when different LLMs are paired with the same GICL method across various diverse datasets or tasks?}

\noindent \textbf{Experiment Settings.}
We selected the Cora and Sports datasets to compare the results of three different LLMs presented in Table~\ref{tab:different_llm} for node classification. To reduce testing costs, we randomly sampled 1,000 data points from the test set of each dataset, and for each dataset, we chose one GICL method for evaluation.

\noindent \textbf{\underline{Observation 3.}} 
\textit{Based on the results, it is reasonable to infer that \textbf{more capable LLMs tend to perform better when integrated with GICL for graph reasoning (GR)}. We also anticipate that future large language models will be incorporated into our GICL benchmark, enabling a deeper investigation of their potential in GR tasks.} 
These differences in LLMs' capabilities are reflected in the consistent ranking of results across both datasets in Table~\ref{tab:different_llm}, where GPT-4o outperforms the other models by 1\%-16\% on both datasets. This demonstrates the significant advantage of GPT-4o in handling graph-related reasoning tasks more effectively. Such performance highlights its superior ability to generalize across varying datasets compared to other competing LLMs.

\subsubsection{The Impact of Structural Information}

\label{sec:4.3}
\textbf{RQ 4.}
\textit{How critical is structural information in graph reasoning tasks?}

\noindent \textbf{Experiment Settings.}
We conducted evaluations on both node classification and link prediction tasks, selecting the most popular dataset for each task as shown in Table~\ref{tab:strcture_matters}. The test sets for both datasets were the same as those described in Section~\ref{sec:4.1}. We employed LLaMA3-70b-Instruct as the backbone for our GraphICL framework.

\noindent \textbf{\underline{Observation 4.}}
\textit{\textbf{Structural information via GraphICL significantly boosts performance}, particularly in datasets where neighbor relationships play a crucial role, such as citation networks.} As illustrated in Table~\ref{tab:strcture_matters}, graph prompts that incorporate structure-aware neighbor information consistently yield better results compared to those that omit such information. This trend is observed across both graph reasoning tasks, highlighting the importance of leveraging structural context in improving performance. Furthermore, the degree of improvement varies depending on whether 1-hop or 2-hop neighbors are selected, with different levels of structural depth contributing uniquely to the reasoning process. These findings underscore the value of integrating graph structure into prompts, enabling models to capture richer contextual relationships and make more informed predictions.



\begin{table}[htbp]
    \caption{Comparison of results with and without structure-aware information. For GraphICL, we adopted the abbreviation format(shown in GraphICL column) as presented in Table~\ref{tab:different_llm}, with further details available in the Appendix~\ref{sec:A.2}.}
    \scriptsize
    \label{tab:strcture_matters}
    \begin{tabularx}{\linewidth}{c X X X X X}
        \toprule
        Dataset & Task & GraphICL & Acc(\%) & $\uparrow_{1-hop}$ & $\uparrow_{2-hop}$\\ 
        \midrule
        \multirow{3}{*}{Arxiv} & \multirow{3}{*}{NC} & XXGR & 30.36 & \multirow{3}{*}{+107.83\%} & \multirow{3}{*}{+111.85\%} \\
        & & 1SGR & 63.10 & &\\
        & & 2SGR & 64.32 & & \\
        \midrule
        \multirow{3}{*}{Cora} & \multirow{3}{*}{LP} & XXCR & 71.91 & \multirow{3}{*}{+4.29\%} & \multirow{3}{*}{+10.83\%} \\
        & & 1RCR & 75.00 & & \\
        & & 2RCR & 79.70 & & \\
        \bottomrule
    \end{tabularx}
    \vspace{-0.4cm}
\end{table}

\subsubsection{Further Analysis}
\noindent \textbf{Similar Neighbors Boost Node Classification.} GraphICL provides three neighbor selection strategies: random, pagerank-based, and similarity-based. To evaluate their effectiveness, we employed LLaMA3-70B-Instruct across four diverse datasets and calculated the average accuracy achieved by each method. As presented in Table~\ref{tab:table4}, the similarity-based method consistently delivered the highest accuracy among the three strategies. This superior performance can be attributed to its focus on text similarity, which effectively identifies neighbors with similar content that often share the same labels. This alignment allows LLMs to extract and leverage meaningful textual cues, facilitating more precise predictions and improving reasoning capabilities within graph-based tasks. Furthermore, by emphasizing content-related connections, the similarity-based approach ensures that the model considers the most relevant information, enhancing its ability to generalize across datasets and scenarios.

\noindent \textbf{Chain-of-Thought: Not Always Beneficial.} We incorporated Chain-of-Thought (CoT) prompting into GraphICL by appending "Let's think step by step" to the prompt~\citep{wei2022chain}, intending to improve the model's reasoning capabilities by guiding it through a structured thought process. However, the results indicate that the impact of CoT is inconsistent and varies depending on the specific method employed. As shown in Table~\ref{tab:table5}, for the 1RGR template, CoT led to a notable improvement in accuracy, increasing it from 75.46\% to 78.41\% (+2.95\%), highlighting its potential to enhance reasoning in certain scenarios. In contrast, the 1SCR template experienced a significant decline in performance, with accuracy dropping from 70.85\% to 67.16\% (-3.69\%) when CoT was applied. When considering the overall results across all 55 settings, the average accuracy without CoT was 65.53\%, while with CoT it decreased slightly to 65.10\% (-0.43\%). These findings suggest that while CoT prompting is not universally effective and may even hinder performance in other cases. This variability underscores the importance of understanding task-specific and method-specific dynamics when integrating CoT strategies into graph-related tasks.

\begin{table}[H]
    \footnotesize
    \caption{The average accuracy of different neighbor selection mechanisms across four datasets.}
    \begin{tabular}{c c c c c}
        \toprule
        Mechanism & Cora & PubMed & Photo & History \\
        \midrule
        Random & 68.45 & 67.25 & 50.30 & 40.13 \\
        Pagerank & 68.40 & 67.16 & 46.66 & 38.10 \\
        Similarity & \textbf{68.76} & \textbf{67.63} & \textbf{59.06} & \textbf{41.97} \\
        \bottomrule
    \end{tabular}
    \vspace{-0.4cm}
    \label{tab:table4}
\end{table}

\begin{table}[H]
    \centering
    \footnotesize
    \caption{Accuracy comparison of different GraphICL Methods on Cora with and without CoT. "Average" represents the mean accuracy of all 55 settings.}
    \begin{tabular}{c c c c}
        \toprule
        GraphICL & CoT & Acc(\%) & $\uparrow_{CoT}$\\
        \hline
        1RGR & No & 75.46 & \multirow{2}{*}{+2.95\%}\\
        1RGR & Yes & 78.41 \\
        \hline
        1SCR & No & 70.85 & \multirow{2}{*}{-3.69\%}\\
        1SCR & Yes & 67.16 \\
        \hline
        Average & No & 65.53 & \multirow{2}{*}{-0.43\%}\\
        Average & Yes & 65.10 \\
        \bottomrule
    \end{tabular}
    \label{tab:table5}
\end{table}

\section{Conclusions}

We introduce GraphICL, a comprehensive and versatile prompt benchmark designed for graph in-context learning using LLMs across a diverse range of graph inference tasks. Through extensive experimental evaluations, we demonstrate that GraphICL empowers LLMs to achieve exceptional performance across multiple datasets, often surpassing state-of-the-art supervised GNNs and specialized graph LLMs in various scenarios. These results highlight the potential of in-context learning to advance graph reasoning. Looking ahead, we aim to expand our benchmark by incorporating additional LLMs and extending the scope of graph-related tasks, with the goal of pushing the boundaries of LLM capabilities in tackling increasingly complex and nuanced graph-based challenges.
\section{Limitation}
We introduce GraphICL, which leverages graph in-context learning to enhance the performance of LLMs in graph reasoning. In terms of breadth, we acknowledge the need to test our template on more classic graph tasks. Additionally, to expand our benchmark, incorporating more large language models is essential for further enrichment. As for depth, given the complexity of graph structures, we need to explore how to better integrate structural information with demonstrations in the prompts, especially for text graphs of varying natures, such as molecular graphs, to achieve better results.

\bibliography{main}

\clearpage
\appendix

\section{Appendix}
\label{sec:appendix}

\subsection{General Prompt Template}
The prompt inputted into LLMs consists of a system prompt, user content, and some special characters. In this section, we will showcase the system prompts and user content we designed for various tasks and datasets.

\textbf{System Prompt Design.}
The system prompt is often used to make the LLMs aware of the task they are about to perform. Table~\ref{tabel6} presents the system prompts used for node classification tasks across different datasets, while Table~\ref{table7} shows the system prompts for link prediction tasks in two major types of datasets.

\textbf{User Content Design.}
In GraphICL, user content is used to record information other than the task description, such as structure-aware text information, anchor node text information, and demonstrations. The specific templates are shown in Table~\ref{table7_a} and Table \ref{table8}.

\begin{figure*}[h]
    \centering
    \caption{Examples of graph in-context learning prompting in different graph reasoning tasks.}
    \label{figure2}
    \includegraphics[width=0.98\linewidth]{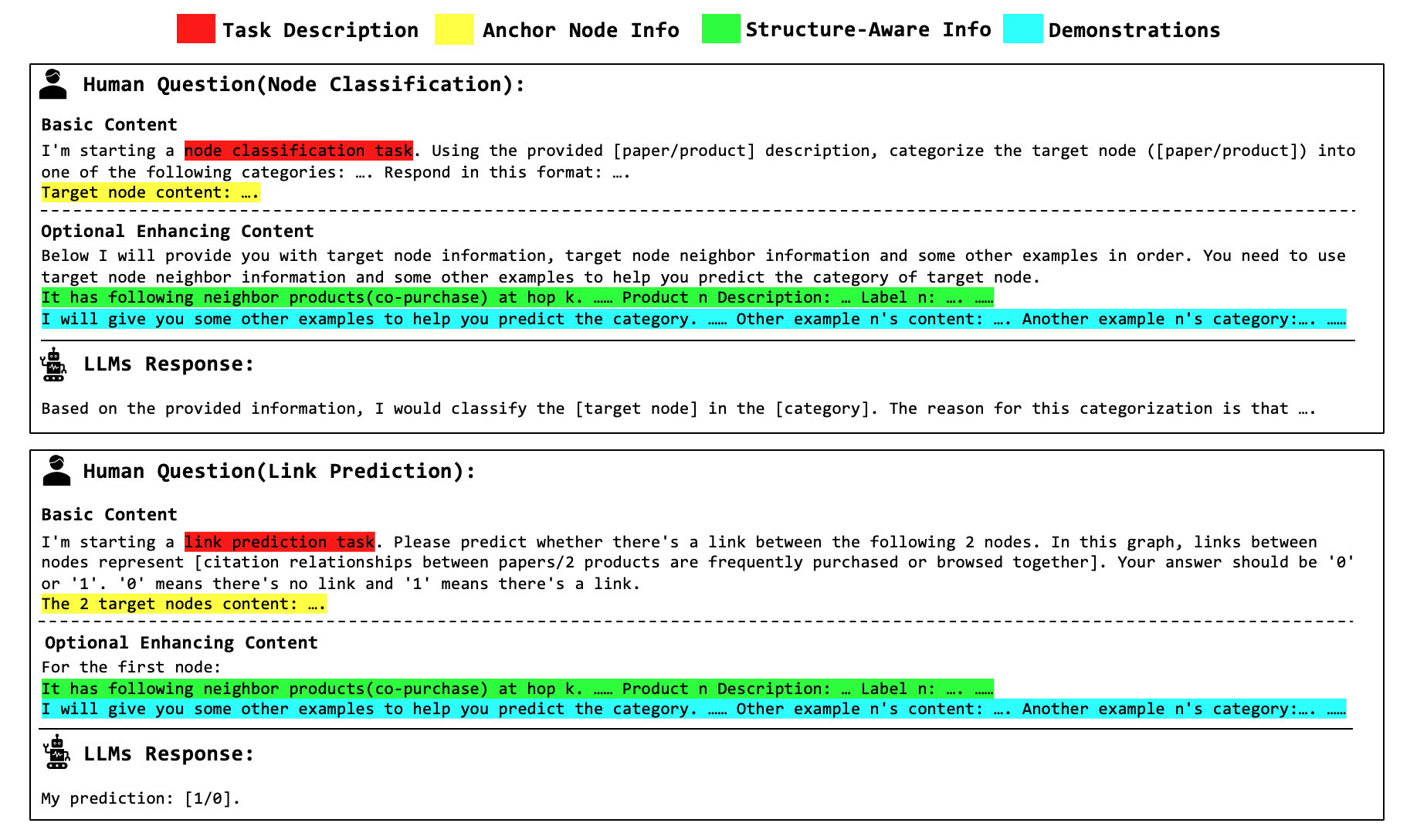}
\end{figure*}

\begin{table*}[h]
    \caption{The GICL settings of in-domain node classification results.}
    \label{tab:GICL_settings of table 1}
    \centering
    \scriptsize
    \renewcommand{\arraystretch}{1.4} 
    \begin{tabularx}{\textwidth}{c X X X X X X}
        \hline
        \rowcolor{blue!10} \textbf{Model} & \textbf{Computers} & \textbf{Sports} & \textbf{PubMed} & \textbf{Cora} & \textbf{Arxiv} & \textbf{Products} \\
        \hline
        GraphICL-LLaMA3-S1 & 1RCP & 1RGP & 2SCS & 1RGR & 1RCP & 1RGS \\
        GraphICL-LLaMA3-S2 & 1RGP & 1RGS & 2SCR & 1SCP & 1RGR & 1RGR \\
        GraphICL-LLaMA2-S1 & 1'SXX & 2SXX & 2'SXX & 2SGR & 2SCP & 2SCP \\
        GraphICL-LLaMA2-S2 & 1'RXX & 1SGS & 2'RXX & 1SGR & 1SCP & 1'SXX \\
        \arrayrulecolor{black}
        \hline
    \end{tabularx}
\end{table*}

\subsection{Methodology}
\label{sec:A.2}
For the zero-shot setting, we first provide the text information of the anchor nodes and implement seven different structure-aware methods: "XX," "1R," "1P," "1S," "2R," "2P," and "2S." In this context, "1" and "2" represent one-hop and two-hop neighbor information, respectively, which is incorporated into the structure-aware content.

In the few-shot setting, there are multiple approaches to implementing demonstrations. In the structure-aware configuration, six methods are used: "1'R," "1'P," "1'S," "2'R," "2'P," and "2'S." Here, "1'" and "2'" indicate that one-hop and two-hop neighbors are used as demonstrations. In the non-structure-aware configuration, seven methods are applied: "GR," "GP," "GS," "CR," "CP," "CS," and "XX."

The absence of "XX" in the structure-aware category is due to the fact that "XX" in the structure-aware context is equivalent to "XX" in the non-structure-aware context. Therefore, "XX" is counted only in the non-structure-aware group. Additionally, "G" and "C" refer to the demonstration selection scope: "G" indicates that demonstrations are selected without regard to labels, while "C" ensures that one demonstration is selected per label from the training set. The letters "R," "P," and "S" indicate the selection mechanisms—random, PageRank, and similarity, respectively.

In total, there are 55 possible combinations: 7 structure-aware methods combined with 7 demonstrations, plus 1 structure-aware "XX" combined with 6 non-structure-aware demonstrations, resulting in $7 \times 7 + 1 \times 6 = 55$ methods.

For the GICL settings we used in Table~\ref{tab:Table1} and Table~\ref{tab:out_graph}, please refer to Table~\ref{tab:GICL_settings of table 1}. The GICL setting we used in the link prediction test (Table ~\ref{tab:link_prediction}) is "1SXX".

\begin{table}[H]
\centering
\small
    \caption{Statistics of the TAG datasets.}
    \label{tab:tag_datasets_stats}
    \begin{tabular}{llll}
    \toprule[1.5pt]
        Dataset & \#Nodes & \#Edges & \#Classes  \\ \hline
        Cora & 2,708 & 5,429 & 7  \\ 
        PubMed & 19,717 & 44,338 & 3 \\
        OGB-Arxiv & 169,343 & 1,166,245 & 40  \\ 
        OGB-Products (subset) & 54,025 & 74,420 & 47  \\ 
        Amazon-Sports & 173,055 & 1,946,555 & 13  \\ 
        Amazon-Computers & 87,229 & 721,107 & 10  \\ 
        Amazon-Photo & 48,362 & 500,939 & 12 \\ 
        Amazon-Children & 76,875 & 1,631,453 & 24  \\ 
        Amazon-History & 41,551 & 358,574 & 13  \\ 
    \bottomrule[1.5pt]
    \vspace{-0.6cm}
    \end{tabular}

\end{table}

\subsection{Experiments}
\label{sec:A.3}
\subsubsection{Evaluation Datasets}
\label{sec:A.3.1}

The statistics for all TAG datasets used in this study can be found in Table~\ref{tab:tag_datasets_stats}. In our node classification experiments, data splitting was rigorously conducted according to established protocols to ensure consistency and comparability of the results. For the Cora, PubMed and OGB-Products datasets, we followed the splits specified by TAPE \citep{he2023harnessing}. For OGB-Arxiv dataset, we used the standard split provided by the OGB framework \citep{hu2020open}, ensuring strict compliance with the benchmark's guidelines. For the other Amazon datasets, we applied a 6:2:2 ratio for training, validation, and testing sets. 

In the supervised setting, the splits for Cora and PubMed were based on TAPE’s guidelines. For OGB-Products, we sampled 5000 instances from the testing set based on the TAPE split. Similarly, we also sampled 5000 instances from the standard testing set. For other Amazon datasets, we followed the 6:2:2 split strategy.

In the semi-supervised setting, for Cora and PubMed, we adopted the standard semi-supervised splits \citep{wang2024cluster}, while for OGB-Products, we applied a 20-shot split. For all Amazon datasets, a 300-shot split was used. Additionally, we ensured that the testing sets in the semi-supervised setting were consistent with those in the supervised setting.

For the link prediction evaluation, in the supervised setting, we followed the same splits as used in LLaGA \citep{chen2024llaga}. In the semi-supervised setting, we randomly sampled 5\% of the examples from the supervised training set, ensuring an equal number of positive and negative samples, while keeping the test set unchanged.

\subsubsection{Computing Environment and Resources}
We leveraged the vLLM package \citep{kwon2023efficient} for inference of large language models. Locally, we deployed the LLaMA2-13B-Chat model on a single NVIDIA A100 80GB GPU and the LLaMA3-70b-Instruct model on two of these GPUs to accommodate its greater computational requirements. For GPT-4o inference, we utilized the OpenAI API.

\subsubsection{Number of Neighbors and Examples}
There is a length constraint on the LLMs' input window. Within this constraint, we determined that a maximum of 6 neighbors or demonstrations can be included in node classfication. In link prediction, we select one of the nodes and provide information about up to six of its neighbors, along with three additional demonstrations (if available). We utilize GIA \citep{chien2021node} embeddings to compute similarity. 

\subsubsection{Node Classification Results}
\label{sec:A.4}
In Table~\ref{tab:Table1}, we report results for the supervised setting of GraphGPT, as the available checkpoints only support joint supervised training on the PubMed and OGB-Arxiv datasets, preventing us from evaluating its semi-supervised performance. And for Table ~\ref{tab:link_prediction}, the 
checkpoint for semi-supervised training on cora is also missing. Similarly, GraphTranslator's self-supervised training does not involve label information from the dataset, making it unsuitable for division into supervised or semi-supervised categories. Therefore, we include its results both under the semi-supervised and supervised setting. For all datasets listed in Table~\ref{tab:Table1}, in addition to using BERT embeddings for MLP and GNN models, we also employed BoW \citep{harris1954distributional} and GIA \citep{chien2021node} embeddings in both semi-supervised and supervised settings, as shown in Table~\ref{tab:extra in-domain results}. 

For LLaGA, GraphPrompter, GraphTranslator, and GraphGPT, we utilized Vicuna-7b-v1.5-16k, LLaMA2-13b-Chat-HF, ChatGLM6B, and Vicuna-7B-v1.5 as their respective LLM backbones. Additionally, for LLaGA, GraphTranslator, and GraphGPT, we used the same types of embeddings as in the original works, while for GraphPrompter, we tested using GIA embeddings \citep{chen2024exploring}. In the in-domain experiments, we adopted a 'single focus' training approach for all models except GraphGPT, meaning that each model was trained on an individual training set and tested on its corresponding test set. For GraphGPT, we directly loaded the model parameters from its mixed training setup for testing. In the cross-domain experiments, all models were trained on multiple training sets jointly. Apart from LLaGA, which followed its original mixed training strategy, other models combined training sets at a 1:1 ratio.

For the three Amazon review datasets (Amazon-Photo, Amazon-Children, and Amazon-History), which are not included in {Table~\ref{tab:Table1}}, we provide both in-domain and cross-domain results in Table~\ref{tab:in-domain results of 3 amazon datasets} and {Table~\ref{tab:extra_cross_domain}}, respectively.

\begin{figure*}[p]
    \centering
    \includegraphics[width=1\linewidth]{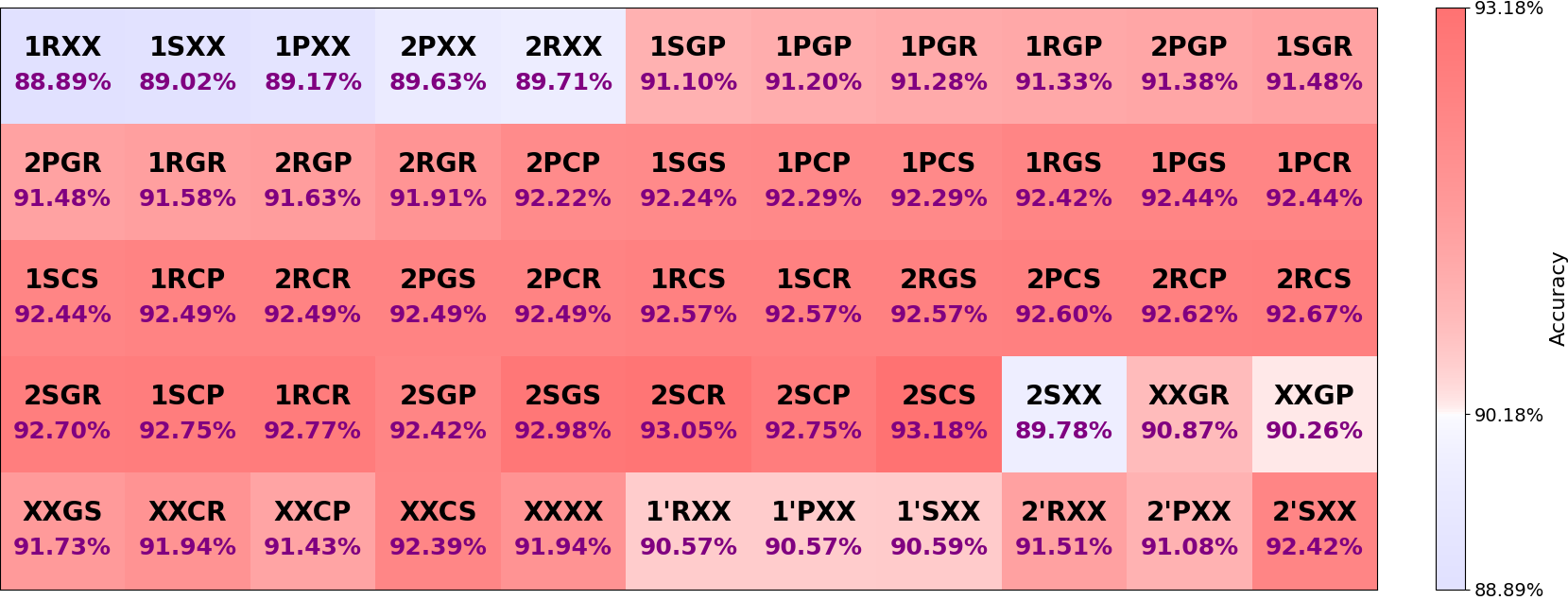}
    \caption{The comparison heat map between the GraphICL method based on LLaMA3-70B-Instruct and the best Supervised GNNs results (SAGE-90.18\%) in the table~\ref{tab:extra in-domain results} in PubMed.  The results of the GNNs are used as the baseline, with higher performance relative to the baseline shown in red and lower performance shown in blue.}
    \label{fig:heatmap}
    \vspace{-0.4cm}
\end{figure*}

We selected PubMed to showcase the best results from the 55 configurations tested with our GraphICL combined with LLaMA3-70B-Instruct, comparing them to supervised GNN methods. Detailed comparisons can be found in Figure~\ref{fig:heatmap}.

\begin{table*}[htbp]
\centering
\small
\begin{tabular}{cccccccc}
\toprule[1.5pt] 
\renewcommand{\arraystretch}{1.4} 
\textbf{Method} & \textbf{Embedding} & \textbf{Computers} & \textbf{Sports} & \textbf{PubMed} & \textbf{Cora} & \textbf{Arxiv} & \textbf{Products} \\
\hline 
\multicolumn{8}{>{\columncolor{gray!20}}c}{\textbf{Semi-Supervised In-Domain Results}} \\
\hline
\arrayrulecolor{black}
\multirow{3}{*}{MLP}   & BoW & 49.69 & 67.17 & 67.14 & 52.95 & 27.38 & 56.80 \\
                       & BERT & 44.56 & 58.74 & 59.38 & 47.23 & 37.10 & 65.36 \\
                       & GIA  & 66.80 & 81.80 & 74.82 & 64.02 & 48.39 & 70.23 \\
\hline
\multirow{3}{*}{GCN}   & BoW & 72.58 & 64.76 & 80.07 & 74.58 & 50.31 & 71.65 \\
                       & BERT & 70.24 & 59.12 & 59.38 & 68.82 & 55.27 & 74.47 \\
                       & GIA  & 81.40 & 76.80 & 77.62 & 69.45 & 51.36 & 74.95 \\
\hline
\multirow{3}{*}{SAGE}  & BoW & 73.16 & 63.23 & 77.72 & 67.23 & 45.73 & 69.21 \\
                       & BERT & 69.53 & 58.52 & 64.66 & 64.58 & 54.05 & 72.35 \\
                       & GIA  & 83.18 & 76.36 & 76.06 & 70.85 & 55.20 & 73.66 \\
\hline
\multirow{3}{*}{RevGAT}    & BoW & 69.05 & 59.41 & 73.28 & 71.40 & 39.41 & 67.99 \\
                           & BERT & 64.63 & 55.48 & 64.10 & 65.31 & 48.86 & 71.45 \\
                           & GIA  & 81.55 & 74.78 & 78.09 & 72.88 & 50.94 & 73.78 \\
\hline
\multicolumn{8}{>{\columncolor{gray!20}}c}{\textbf{Supervised In-Domain Results}} \\
\hline
\arrayrulecolor{black}
\multirow{3}{*}{MLP}   & BoW & 64.90 & 84.12 & 71.88 & 74.72 & 55.59 & 58.83 \\
                       & BERT & 61.74 & 85.58 & 82.28 & 60.89 & 66.07 & 67.56 \\
                       & GIA  & 75.72 & 90.97 & 90.04 & 77.12 & 71.64 & 70.91 \\
\hline
\multirow{3}{*}{GCN}   & BoW & 77.99 & 88.87 & 86.76 & 88.19 & 71.31 & 72.06 \\
                       & BERT & 74.84 & 88.33 & 85.51 & 86.90 & 72.82 & 75.56 \\
                       & GIA  & 82.74 & 91.97 & 88.82 & 88.39 & 73.56 & 75.36 \\
\hline
\multirow{3}{*}{SAGE}  & BoW & 76.86 & 88.81 & 85.46 & 88.93 & 70.43 & 70.25 \\
                       & BERT & 73.62 & 89.21 & 85.66 & 83.39 & 72.54 & 73.44 \\
                       & GIA  & 82.98 & 92.64 & 90.18 & 82.95 & 74.07 & 74.38 \\
\hline
\multirow{3}{*}{RevGAT}& BoW & 77.91 & 89.76 & 89.56 & 86.90 & 70.59 & 70.93 \\
                       & BERT & 72.87 & 88.70 & 86.01 & 82.66 & 73.21 & 73.62 \\
                       & GIA  & 83.43 & 92.94 & 88.92 & 82.47 & 74.74 & 74.88 \\
\hline
\multicolumn{8}{>{\columncolor{gray!20}}c}{\textbf{Results of GraphICL}} \\
\hline
\arrayrulecolor{black}
GraphICL-LLaMA3-S1 & -  & \textbf{87.37} & \textbf{91.39} & \textbf{93.18} & \textbf{83.58} & \textbf{73.68} & \textbf{81.48} \\
                          GraphICL-LLaMA3-S2 & -  & 87.37 & 91.12 & 93.05 & 83.21 & 73.54 & 81.04 \\
\bottomrule[1.5pt]
\end{tabular}
\caption{Extended in-domain Results from \textbf{Table~\ref{tab:Table1}} using different embedding types for MLP and GNN models.}
    \label{tab:extra in-domain results}
    \vspace{-0.4cm}
\end{table*}

\begin{table*}[htbp]
\centering
\begin{tabular}{ccccc}
\toprule[1.5pt] 
\renewcommand{\arraystretch}{1.4} 
\textbf{Method} & \textbf{Embedding} & \textbf{Photo} & \textbf{Children} & \textbf{History}\\
\hline 
\multicolumn{5}{>{\columncolor{gray!20}}c}{\textbf{Semi-Supervised In-Domain Results}} \\
\hline
\arrayrulecolor{black}
\multirow{3}{*}{MLP}   & BoW & 51.07 & 25.96 & 58.47 \\
                       & BERT & 42.08 & 31.54 & 69.41  \\
                       & GIA  & 66.70 & 36.70 & 74.21 \\
\hline
\multirow{3}{*}{GCN}   & BoW & 63.05 & 30.77 & 64.60  \\
                       & BERT & 59.78 & 35.06 & 69.51  \\
                       & GIA  & 69.80 & 34.13 & 71.67 \\
\hline
\multirow{3}{*}{SAGE}  & BoW & 63.20 & 30.84 & 68.09  \\
                       & BERT & 59.75 & 35.61 & 73.79 \\
                       & GIA  & 71.44 & 40.01 & 75.75\\
\hline
\multirow{3}{*}{RevGAT}    & BoW & 60.03 & 29.97 & 61.41  \\
                           & BERT & 54.07 & 34.60 & 70.99 \\
                           & GIA  & 70.08 & 36.96 & 73.22 \\
\hline
\multicolumn{5}{>{\columncolor{gray!20}}c}{\textbf{Supervised In-Domain Results}} \\
\hline
\arrayrulecolor{black}
\multirow{3}{*}{MLP}   & BoW & 68.50 & 49.71 & 77.37\\
                       & BERT & 67.93 & 51.46 & 82.41 \\
                       & GIA  & 79.73 & 55.96 & 84.13\\
\hline
\multirow{3}{*}{GCN}   & BoW & 77.05 & 53.56 & 81.12\\
                       & BERT & 77.08 & 54.53 & 83.45\\
                       & GIA  & 82.62 & 55.23 & 84.27\\
\hline
\multirow{3}{*}{SAGE}  & BoW & 77.41 & 54.86 & 80.82\\
                       & BERT & 76.40 & 55.27 & 84.06 \\
                       & GIA  & 83.28 & 58.41 & 85.12\\
\hline
\multirow{3}{*}{RevGAT}& BoW & 77.84 & 52.96 & 80.97\\
                       & BERT & 75.87 & 53.10 & 83.09\\
                       & GIA  & 83.33 & 55.73 & 84.38\\

\hline
\multicolumn{5}{>{\columncolor{gray!20}}c}{\textbf{Results of GraphICL}} \\
\hline
\arrayrulecolor{black}
GraphICL-LLaMA3-S1 & - & \textbf{79.35} & \textbf{47.96} & \textbf{80.89} \\
GraphICL-LLaMA3-S2 & - & 77.78 & 47.63 & 79.18 \\
\bottomrule[1.5pt]
\end{tabular}
\caption{In-domain results of amazon-photo, amazon-children, and amazon-history. For the Amazon-Photo dataset, S1 is "1RGS" and S2 is "1RCS". For Amazon-History, S1 is "1SGS" and S2 is "1PGS". For Amazon-Children, S1 is "1RGP" and S2 is "1RGS".}
    \label{tab:in-domain results of 3 amazon datasets}
\end{table*}

\begin{table*}[htbp]
    \centering
    \renewcommand{\arraystretch}{1.2} 
 
    \begin{tabular}{ccccc}
        \toprule[1.5pt]
        \textbf{Method} & \textbf{Photo} & \textbf{Children} & \textbf{History} \\
        \hline
        \multicolumn{4}{>{\columncolor{gray!20}}c}{\textbf{Supervised Cross-Domain Results (GraphLLMs)}} \\
        \hline
        LLaGA-ND & 19.83 & 7.49 & 6.45 \\
        LLaGA-HO & 6.16 & 11.14 & 7.94 \\
        GraphGPT & 6.18 & 14.56 & 10.94 \\
        GraphPrompter & 25.01 & 10.35 & 14.62 \\
        GraphTranslator & 38.96 & 16.13 & 6.64 \\
        \hline
    
        \multicolumn{4}{>{\columncolor{gray!20}}c}{\textbf{Results of GraphICL}} \\
        \hline
        \arrayrulecolor{orange}
        GraphICL-LLaMA3-S1 & \textbf{79.35} & \textbf{47.96} & \textbf{80.89} \\
        GraphICL-LLaMA3-S2 & 77.78 & 47.63 & 79.18 \\
        \arrayrulecolor{black}
    \bottomrule[1.5pt]
    \end{tabular}
    \caption{Cross-Domain Results of Amazon-Photo, Amazon-Children, and Amazon-History. S1 and S2 are the same as Table~\ref{tab:in-domain results of 3 amazon datasets}}
    \label{tab:extra_cross_domain}
\end{table*}

\begin{table*}[htbp]
\centering
\footnotesize
\begin{tabularx}{\textwidth}{>{\hsize=.67\hsize}X >{\hsize=1.33\hsize}X}
\toprule[1.5pt]
\textbf{Dataset} & \textbf{System Prompt Content} \\ \hline
Cora & I'm starting a node classification task. Please predict the most appropriate category for the target node (paper). Choose from the following categories: \textcolor{gray}{\textbackslash n} Rule Learning \textcolor{gray}{\textbackslash n} Neural Networks \textcolor{gray}{\textbackslash n} Case Based \textcolor{gray}{\textbackslash n} Genetic Algorithms \textcolor{gray}{\textbackslash n} Theory \textcolor{gray}{\textbackslash n} Reinforcement Learning \textcolor{gray}{\textbackslash n} Probabilistic Methods. \\ \midrule
PubMed & I'm starting a node classification task. Please predict the most likely type of the target node (paper). Your answer should be chosen from: \textcolor{gray}{\textbackslash n} Type 1 diabetes. \textcolor{gray}{\textbackslash n} Type 2 diabetes. \textcolor{gray}{\textbackslash n} Experimentally induced diabetes. \\ \midrule
OGB-Arxiv & I'm starting a node classification task. Please predict the most appropriate Arxiv Computer Science (CS) sub-category for the target node (paper). The predicted sub-category should be in the format 'cs.XX'. \\ \midrule
Amazon-History & I'm starting a node classification task. Using the provided history-related book's title and description, categorize the target node (book) into one of the following categories: ['Americas', 'Asia', 'Australia \& Oceania', 'World', 'Europe', 'Middle East', 'Historical Study \& Educational Resources', 'Arctic \& Antarctica', 'Ancient Civilizations', 'Africa', 'Russia', 'Military']. Respond in this format: The book belongs to the [Category] category due to [evidence from the book product descriptions]. \\ \midrule
Amazon-Computers & I'm starting a node classification task. Given the product review provided, please categorize the target node (product) into one of the following categories: ['Tablet Replacement Parts', 'Monitors', 'Networking Products', 'Computers \& Tablets', 'Computer Accessories \& Peripherals', 'Tablet Accessories', 'Laptop Accessories', 'Computer Components', 'Data Storage', 'Servers']. Your classification should be based on the content of the review. Please support your answer with evidence from the review. Response Format: The product falls under the category of [Category]. This determination is based on the product review, where [specific details from the review supporting the classification]. \\ \midrule
Amazon-Photo & I'm starting a node classification task. Given the product review provided, please categorize the target node (product) into one of the following categories: ['Flashes', 'Film Photography', 'Accessories', 'Lighting \& Studio', 'Video Surveillance', 'Underwater Photography', 'Digital Cameras', 'Tripods \& Monopods', 'Lenses', 'Video', 'Binoculars \& Scopes', 'Bags \& Cases'] Your classification should be based on the content of the review. Please support your answer with evidence from the review. Response Format: The product falls under the category of [Category]. This determination is based on the product review, where [specific details from the review supporting the classification]. \\ \midrule
Amazon-Book & I'm starting a node classification task. Using the provided children book's title and description, categorize the target node (book) into one of the following categories: ['Literature \& Fiction', 'Animals', 'Growing Up \& Facts of Life', 'Humor', 'Cars, Trains \& Things That Go', 'Fairy Tales, Folk Tales \& Myths', 'Activities, Crafts \& Games', 'Science Fiction \& Fantasy', 'Classics', 'Mysteries \& Detectives', 'Action \& Adventure', 'Geography \& Cultures', 'Education \& Reference', 'Arts, Music \& Photography', 'Holidays \& Celebrations', 'Science, Nature \& How It Works', 'Early Learning', 'Biographies', 'History', 'Children's Cookbooks', 'Religions', 'Sports \& Outdoors', 'Comics \& Graphic Novels', 'Computers \& Technology']. Please provide your reasoning. Respond in this format: The book belongs to the [Category] category due to [evidence from the book product descriptions]. \\ \midrule
Amazon-Sports & I'm starting a node classification task. Using the provided item's title in the Sports \& Fitness category, categorize the target node (item) into one of the following categories: ['Other Sports', 'Exercise \& Fitness', 'Hunting \& Fishing', 'Accessories', 'Leisure Sports \& Game Room', 'Team Sports', 'Boating \& Sailing', 'Swimming', 'Tennis \& Racquet Sports', 'Golf', 'Airsoft \& Paintball', 'Clothing', 'Sports Medicine']. Please provide your reasoning. Respond in this format: The item belongs to the [Category] category due to [evidence from the item descriptions]. \\ \midrule
OGB-Products & I'm starting a node classification task. Using the provided amazon product's title and description, please predict the most likely category of this node (product) from Amazon. Your answer should be chosen from the following categories: (Categories omitted due to length) \\ 
\bottomrule[1.5pt]
\end{tabularx}
\caption{System prompts for node classification tasks across various datasets.}
\label{tabel6}
\end{table*}

\begin{table*}[htbp]
\centering
\begin{tabularx}{\textwidth}{>{\hsize=.67\hsize}X >{\hsize=1.33\hsize}X}
\toprule[1.5pt]
\textbf{Dataset Type} & \textbf{System Prompt Content} \\ \midrule
Citation Network & I'm starting a link prediction task. Please predict whether there's a link between the following 2 nodes. In this graph, links between nodes represent the citation relationships between papers. Your answer should be '0' or '1'. '0' means there's no link and '1' means there's a link. \\ \midrule
Amazon Review Dataset & I'm starting a link prediction task. Please predict whether there's a link between the following 2 nodes. In this graph, links between nodes represent that 2 \textcolor{blue}{<specific type>} products are frequently purchased or browsed together. Your answer should be '0' or '1'. '0' means there's no link and '1' means there's a link. \\ \bottomrule[1.5pt]
\end{tabularx}
\caption{System prompts for link prediction tasks across two types of datasets (Citation Networks and Amazon Datasets).}
\label{table7}
\end{table*}

\begin{table*}[htbp]
\centering
\begin{tabularx}{\textwidth}{X X}
\toprule[1.5pt]
\textbf{GraphICL} & \textbf{User Content} \\ \hline
Zero-Shot without Structure-Aware Information 
& Below I will provide you with target node information. (Please reason step by step.) \textcolor{gray}{\textbackslash n} Target node content: \textcolor{blue}{<Target Node Text>}. \\
\midrule
Zero-Shot with Structure-Aware Information  & Below I will provide you with target node information and target node neighbor information. You need to use target node neighbor information to help you predict the category of target node.  (Please reason step by step.) \textcolor{gray}{\textbackslash n} Target node content: \textcolor{blue}{<Target Node Text>}. \textcolor{gray}{\textbackslash n} It has following neighbor \textcolor{blue}{<products(co-purchase) / books / papers>} at hop \textcolor{blue}{<Number of Hops>: [Neighbors' Text>}. \\
\midrule
Few-Shot without Structure-Aware Information & Below I will provide you with target node information and some other examples in order. You need to use examples to help you predict the category of target node.  (Please reason step by step.) \textcolor{gray}{\textbackslash n} Target node content: \textcolor{blue}{<Target Node Text>}. \textcolor{gray}{\textbackslash n} I will give you some other examples to help you predict the category: \textcolor{blue}{<Example's Text, Example's Label>}. \\
\midrule
Few-Shot with Structure-Aware Information & Below I will provide you with target node information, target node neighbor information and some other examples in order. You need to use target node neighbor information and some other examples to help you predict the category of target node. (Please reason step by step.) \textcolor{gray}{\textbackslash n} Target node content: \textcolor{blue}{<Target Node Text>}. \textcolor{gray}{\textbackslash n} It has following neighbor \textcolor{blue}{<products(co-purchase) / books / papers>} at hop \textcolor{blue}{<Number of Hops>: [Neighbors' Text>}. \textcolor{gray}{\textbackslash n} I will give you some other examples to help you predict the category: \textcolor{blue}{<Example's Text, Example's Label>}. \\
\bottomrule[1.5pt]
\end{tabularx}
\caption{User Content for node classification tasks across two types of datasets (Citation Networks and Amazon Datasets).}
\label{table8}
\end{table*}

\begin{table*}[htbp]
\centering
\begin{tabularx}{\textwidth}{X X}
\toprule[1.5pt]
\textbf{GraphICL} & \textbf{User Content} \\ \hline
Zero-Shot without Structure-Aware Information  & Below I will provide you with target 2 nodes information. \textcolor{gray}{\textbackslash n}
The 2 target nodes content: \textcolor{blue}{<2 Target Nodes Text>} \\
\midrule
Zero-Shot with Structure-Aware Information & Below I will provide you with target 2 nodes information and the first node's neighbor information in order. You need to use the first node's neighbor information to help you predict the link between the 2 target nodes. \textcolor{gray}{\textbackslash n} The 2 target nodes content: \textcolor{blue}{<2 Target Nodes Text>} \textcolor{gray}{\textbackslash n} For the first node: It has following neighbor papers at hop \textcolor{blue}{<Number of Hops>}: \textcolor{blue}{<Neighbors' Text>}. \\
\midrule
Few-Shot without Structure-Aware Information & Below I will provide you with target 2 nodes information and some other examples of node pairs and connections in order. You need to use the other examples to help you predict the link between the 2 target nodes. \textcolor{gray}{\textbackslash n} The 2 target nodes content: \textcolor{blue}{<2 Target Nodes Text>}. \textcolor{gray}{\textbackslash n} The following are the some other examples of node pairs and connections: \textcolor{blue}{<Examples of Node Pairs' Text, Connected: Yes/No>} \\
\midrule
Few-Shot with Structure-Aware Information & Below I will provide you with target 2 nodes information, the first node's neighbor information and some other examples of node pairs and connections in order. You need to use the first node's neighbor information and other examples to help you predict the link between the 2 target nodes. \textcolor{gray}{\textbackslash n} The 2 target nodes content: \textcolor{blue}{<2 Target Nodes Text>}. \textcolor{gray}{\textbackslash n} For the first node: It has following neighbor papers at hop \textcolor{blue}{<Number of Hops>}: \textcolor{blue}{<Neighbors' Text>}. \textcolor{gray}{\textbackslash n} The following are the some other examples of node pairs and connections: \textcolor{blue}{<Examples of Node Pairs' Text, Connected: Yes/No>} \\
\bottomrule[1.5pt]
\end{tabularx}
\caption{User Content for link prediction tasks across two types of datasets (Citation Networks and Amazon Datasets).}
\label{table7_a}
\end{table*}

\end{document}